# Reducing Commitment to Tasks with Off-Policy Hierarchical Reinforcement Learning

**Mitchell Keith Bloch**

University of Michigan
2260 Hayward Street
Ann Arbor, MI. 48109-2121
bazald@umich.edu

**Abstract**

In experimenting with off-policy temporal difference (TD) methods in hierarchical reinforcement learning (HRL) systems, we have observed unwanted on-policy learning under reproducible conditions. Here we present modifications to several TD methods that prevent unintentional on-policy learning from occurring. These modifications create a tension between exploration and learning. Traditional TD methods require commitment to finishing subtasks without exploration in order to update Q-values for early actions with high probability. One-step intra-option learning and temporal second difference traces (TSDT) do not suffer from this limitation. We demonstrate that our HRL system is efficient without commitment to completion of subtasks in a cliff-walking domain, contrary to a widespread claim in the literature that it is critical for efficiency of learning. Furthermore, decreasing commitment as exploration progresses is shown to improve both online performance and the resultant policy in the taxicab domain, opening a new avenue for research into when it is more beneficial to continue with the current subtask or to replan.

## 1 Introduction and Background

Hierarchical reinforcement learning (HRL) is an established solution for attacking the curse of dimensionality. Decomposing a problem into a hierarchy of tasks has a number of advantages. Firstly, knowledge about the values of different tasks and how to perform different tasks can be represented and learned independently. Secondly, there is the possibility of increasing state abstraction at each point in the decision process. Thirdly, available actions can be ignored in some subproblems, reducing the complexity of learning individual decisions. Finally, an agent can share subtasks between parts of a problem, allowing an agent to take advantage of regularities in the behavior demanded by the environment.

Off-policy temporal difference (TD) methods allow an agent to learn about a policy that is different than that which is being followed. This allows an agent to learn reliably from a greater variety of exploration strategies.

We have observed that off-policy TD methods can result in inadvertent on-policy updates in the context of HRL. An example is discussed at length in section 2. Specifically, this can occur if learning is attempted in a task while a non-greedy action is being taken in a subtask. As taking non-greedy actions is fundamental to exploration, this raises the question of how best to reliably learn off-policy without requiring subtasks to converge first.

In exploring solutions to this new problem, we challenge the widespread claim that committing to completion of tasks is a critical aspect of what gives HRL an advantage over flat reinforcement learning (Kaelbling 1993; Dietterich 1998; 2000a; Ryan 2004b; 2004a). Section 3 describes our HRL system. Using corrected TD methods, and particularly a gated version of temporal second difference traces (TSDT) (Bloch 2011), we demonstrate that it is possible to learn efficiently with no commitment to completing tasks. These results are presented in section 4.1. Furthermore, we demonstrate that it is possible to learn more efficiently with a reduction in commitment, opening a research question as to when commitment has value. These results are presented in section 4.2.

### 1.1 Hierarchical Reinforcement Learning

Here we briefly discuss Hierarchical Semi-Markov Q-learning (HSMQ), a close relative of the HRL system presented in section 3; all-goals updating, a technique to improve efficiency of learning; and non-hierarchical or polling execution of tasks, a technique for improving the quality of a learned hierarchical policy.

The design of HSMQ (Dietterich 2000b) keeps the goals of tasks truly independent, sacrificing guarantees of achieving hierarchical optimality. Each task is concerned only with achieving its goal as efficiently as possible, assuming an episodic structure to all tasks. Rewards occurring after a task completes do not affect learning within the task. HSMQ instead guarantees convergence to a recursively optimal policy, by which it is meant that the hierarchy can converge to a policy that is the best given the restrictions imposed by the hierarchy.

Kaelbling (1993) introduced all-goals updating, a technique which concurrently improves an agent's knowledge of how to achieve multiple goals, making much better use of the information acquired from the environment. Dietterich (1998) introduced a subset of all-goals updating, all-states updating, in which only goals that the agent is trying to achieve are updated. The changes required to implement all-states updating in an existing hierarchical reinforcement learning agent are much simpler than the changes required to implement all-goals updating but, depending on a number of factors, it may be significantly less powerful. All-goals updating requires the use of off-policy TD methods. Therefore, we view all-goals updating as one motivation for developing correct off-policy TD methods for HRL.

Both Kaelbling (1993) and Dietterich (1998) discuss non-hierarchical or polling execution of tasks. However, they describe them as techniques for improving the performance of learned policies. Actually learning while executing non-hierarchically is not permitted. We demonstrate a system capable of learning while executing non-hierarchically in section 3. Sutton et al. (1998) presents a simple proof for why executing non-hierarchically must result in a policy that is at least as good as the original.

**One-Step Intra-Option Learning** Sutton and Precup (Sutton and Precup 1998; Sutton, Precup, and Singh 1999) introduced one-step intra-option learning. Here we present how backups must be performed for off-policy learning. Intra-option learning is our basis for reliable learning while executing non-hierarchically, and understanding it is critical for understanding the tradeoffs between different algorithms, as described in section 2.

If an action is non-primitive, representing a subtask, and the subtask does not complete, then the backup must use the Q-value for the same action from the successor state. To use a Q-value for a different action would conflate the issues of deciding how to behave, and learning different behaviors.

$$Q(s,a) \xleftarrow{\alpha} r + \gamma Q(s',a) \qquad (1)$$

Provided that the learning rate, $\alpha$, is sufficiently low, and that actions are sampled adequately over time, $Q(s,a)$ should converge to the true value of the expected return for following action $a$ to completion and then behaving optimally from that point on.

If an action is primitive or successfully terminates a corresponding subtask, then the backup must use the Q-value for the highest valued action from the successor state (if the agent is attempting to learn off-policy).

$$Q(s,a) \xleftarrow{\alpha} r + \gamma V(s') \qquad (2)$$

Provided that $\alpha$ is sufficiently low, and that actions are sampled adequately over time, $Q(s,a)$ should converge to the true value of the expected return for executing action $a$ and then behaving optimally from that point on.

Successful termination of the task at hand (rather than simply a subtask) demands special care unless the agent enters an absorbing state. An absorbing state is a state where there are no actions available to the agent, and therefore has an estimated return of 0. If an absorbing state is not guaranteed, the task should ignore any expected future return for the state, taking only the immediate reward into account.

$$Q(s,a) \xleftarrow{\alpha} r \qquad (3)$$

Provided that $\alpha$ is sufficiently low, and that actions are sampled adequately over time, $Q(s,a)$ should converge to the true value of the expected terminal reward for executing action $a$. Of course, the task may not always terminate when the agent takes action $a$ in a non-deterministic domain. This possibility does not affect the reliability of convergence, given an appropriate $\alpha$.

**Temporal Second Difference Traces** Bloch (2011) introduced temporal second difference traces (TSDT), an alternative to Watkins' Q($\lambda$) with a number of advantages. Storing local $\delta$s for intra-option learning, TSDT can do off-policy backups after non-greedy actions have been taken. This can make it significantly more powerful, particularly in deterministic domains.

In doing backups similarly to intra-option learning, and updating earlier backups as an agent explores, TSDT conceptually approximates Dyna-Q with sample backups but without the need for a model.

## 2 Difficulty Learning Off-Policy

Here we present a three-armed bandit problem, designed to demonstrate a basic problem with existing off-policy temporal difference (TD) methods in the context of HRL. Additionally, we present modifications to popular TD methods to ensure correct off-policy updates.

Actions A, B, and C yield 1, 10, and 100 reward respectively. All three actions result in immediate termination of the episode. It is trivial to develop a flat agent to learn the domain, as depicted in figure 1, but developing a hierarchy should not create additional difficulties. However, it turns out that the trivial hierarchical reinforcement learning (HRL) agent depicted in figure 2 causes problems for traditional off-policy temporal difference methods.

The agent depicted in figure 1 simply chooses action A, B, or C and then terminates. Learning with $\alpha = 1$ and exploring with a fixed exploration strategy (for example epsilon-greedy) and choosing a non-greedy action 10% of the time, it is guaranteed to converge to the optimal policy as time goes to infinity.

There is little incentive to develop a hierarchical agent for this domain. Regardless, one would not naively expect a hierarchy to make it more difficult to converge to an optimal policy. However this is exactly what happens with the hierarchy depicted in figure 2 unless special care is taken. Conceptually, it may be helpful to think of the subtask consisting of a choice between actions A and C as a more complex task than action B. The subtask is preferable to action B only if the subtask is executed expertly.

As depicted in figure 3, a naive implementation of Q-learning will cause significantly decreased performance in our bandit problem given the hierarchy in figure 2. If the effect of exploration in the subtask is not accounted for, the root task will regularly mislearn the value of the subtask, effectively doing a somewhat on-policy backup. Action B will be preferred to the superior subtask 50% of the time because the backup in the root node uses the reward received from the subtask rather than the reward it would have received had the subtask executed greedily.

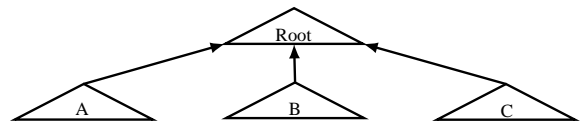

Figure 1: A trivial flat agent for the three-armed bandit problem can choose between actions A, B, and C.

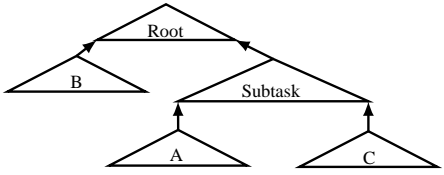

Figure 2: Arbitrary hierarchy which happens to behave interestingly in the toy domain.

A task which is exploring, thereby taking a non-greedy action, is not trying to accomplish its goal. It is trying to gain information. In an off-policy backup, we must then exclude the effects of exploration and block backups that would otherwise occur at higher levels of the hierarchy.

The solution for Watkins' $Q(\lambda)$ (and Q-learning when using all-states updating) is to skip the backup and to just clear the trace whenever a non-greedy action is being taken in any subtask. This enables reliable off-policy learning in HRL systems. Unfortunately, this entails a dilemma as to whether to restrict subtasks to greedy policies while trying to learn at a given level of the hierarchy, or to throw out all potential learning when non-greedy actions are taken at lower levels of the hierarchy. The former option is rather onerous, preventing the agent from learning at all levels of the hierarchy concurrently. The latter option, unfortunately, makes it much less likely that Q-values near the beginning of a subtask will be updated in supertasks if exploration is equally likely throughout the execution of the subtask. A hybrid approach may be possible in which the benefit of one approach could be dynamically weighed against the other, but this could be challenging.

The solution for one-step intra-option learning is simply to skip the backup whenever a non-greedy action is being taken in any subtask. Given the local nature of its backups, this does not suffer from the same dilemma of having to choose between greedy policies for subtasks or the prospect of throwing out potential learning. All that is necessary for convergence is that subtasks not starve their supertasks for updates by constantly choosing non-greedy actions. This happens to be ensured by the standard requirements that action selection be both non-starving and greedy in the limit with infinite experience (GLIE). The downside to one-step intra-option is that learning is relatively slow, taking as many episodes for the agent to learn the value of a task as there are steps in the task. By comparison, $Q(\lambda)$ using all-states updating can learn the value of a task in just one episode under ideal circumstances.

The solution for temporal second difference traces (TSDT) is simply to skip entering the backup into the trace whenever a non-greedy action is being taken in any subtask. These gaps in the traces may reduce the flow of information, but keeping the rest of the trace intact is likely to allow faster learning than one-step intra-option learning so long as there exists the possibility of returning to a previously visited state. Additionally, as backups are done locally as in one-step intra-option learning, TSDT does not suffer from the dilemma we've introduced for $Q(\lambda)$. This makes TSDT ideal for use in our off-policy HRL system as detailed in the following section. Like $Q(\lambda)$, TSDT is capable of propagating reward from the end of a task all the way back to the beginning in a single episode. Reward will propagate back at least as efficiently as when using one-step intra-option learning in the worst case.

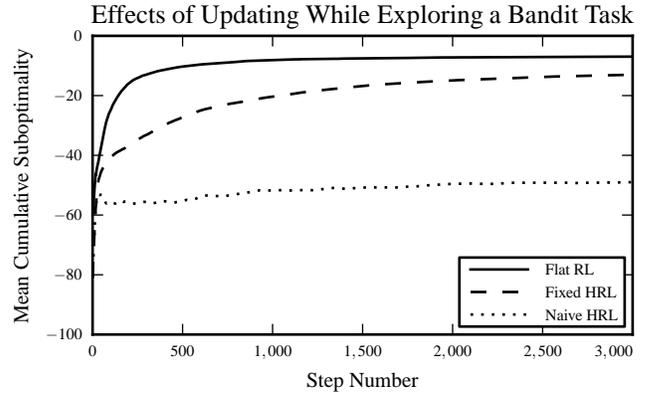

Figure 3: This plot depicts the online performance of agents following a fixed epsilon-greedy exploration strategy in a bandit task. Fixed HRL performs worse than Flat RL only because of the combined effect of exploration in both the root task and the subtask. Naive HRL however has an incorrect policy 50% of the time.

## 3 Off-Policy Hierarchical Reinforcement Learning (OPHRL)

The execution of an agent learning with OPHRL, as described in algorithm 1, takes the form of a traditional hierarchical reinforcement learning algorithm, but executes in a non-hierarchical or polling fashion while learning. At each step of execution, there is no commitment at any level of the hierarchy to continue with the same subtask that was being run in the previous step. This gives one-step intra-option learning and temporal second difference traces (TSDT) certain advantages over Watkins' $Q(\lambda)$, as described in section 2. OPHRL additionally supports arbitrary reward rejection and transformation functions on a per-task basis, allowing solution of the hierarchical credit assignment problem. OPHRL will converge to the true value function for a task regardless of whether exploration is decreased at all, so long as the exploration policy is non-starving for all state-action pairs.

### 3.1 Not Committing to Tasks

Taking big steps of exploration has been cited as a significant advantage of hierarchical reinforcement learning (Dietterich 1998). It has even been suggested that committing to completing all tasks that an agent chooses to begin is necessary for hierarchical reinforcement learning to offer advantages over flat reinforcement learning (Ryan 2004a). It is worth

**Algorithm 1** Off-Policy Hierarchical Reinforcement Learning (OPHRL) using one-step intra-option learning. OPHRL(Root) is called each step. Functions rejectReward and transformReward are task-specific.

**Ensure:** $Q$ initialized arbitrarily,
  e.g., $Q_i(s, a) = 0, \forall$ tasks $i, \forall s \in \mathcal{S}_i^+, \forall a \in \mathcal{A}_i(s)$
1: **function** OPHRL(Task $i$)
2:   Observe $s$
3:   Choose $a$ from $\mathcal{A}_i(s)$ {non-starving}
4:   **if** isTask($a$) **then**
5:     $r, s'$, exploringInSubtask $\Leftarrow$ OPHRL($a$)
6:   **else**
7:     Take primitive action $a$
8:     Observe reward, $r$, and next state, $s'$
9:   **end if**
10:  **if not** exploringInSubtask
     **and not** rejectReward(Task $i$, ...) **then**
11:    $r' \Leftarrow$ transformReward(Task $i$, ...)
12:    **if** $s' \notin \mathcal{S}_i^+$ **then** {Completed task $i$}
13:      $Q_i(s, a) \xleftarrow{\alpha} r'$
14:    **else if** $i \in \mathcal{A}(s')$ **then** {Task $i$ can continue}
15:      **if** isTask($a$) **and** $a \in \mathcal{A}(s')$ **then** {Can continue $a$}
16:        $Q_i(s, a) \xleftarrow{\alpha} r' + \gamma Q_i(s', a)$
17:      **else** {Completed subtask/primitive $a$}
18:        $Q_i(s, a) \xleftarrow{\alpha} r' + \gamma V_i(s')$
19:      **end if**
20:    **end if**
21:  **end if**
22:  **return** $r, s', [Q_i(s, a) < V_i(s)$ **or** exploringInSubtask$]$
23: **end function** OPHRL

noting that Ryan (2004a) acknowledges that it is conceivable that an algorithm without a requirement of commitment could be developed. OPHRL avoids this commitment and, in doing so, has a rather different structure than previous hierarchical reinforcement learning algorithms.

Though commitment to tasks is not an integral part of OPHRL, it is trivial to modify OPHRL to support commitment to tasks. In fact, there is likely some value in doing so. However, folding the question of whether big steps of exploration should be taken into the dilemma of exploration versus exploitation is preferable to restricting agents to big steps. It is well known that an agent can perform better when it is not forced to complete tasks it begins (Sutton, Precup, and Singh 1999). Furthermore, as agents can choose to continue with a subtask until completion, an agent that can abandon a subtask before it completes is guaranteed to be able to explore at least as effectively as an agent that cannot. Whether it is possible to do better, in general, is a difficult question.

### 3.2 Credit Assignment Problem

Cases exist in which rewards do not apply to certain levels of the hierarchy. It could be that a subtask hasn't learned that certain actions are only legal in a subset of the state space. It could be that a supertask misplanned and a large negative reward is received due to no fault of the given task.

Dietterich (1998) addressed the hierarchical credit assignment problem by transforming the reward function. However, these rewards must be rejected outright when using one-step intra-option learning or TSDT. A 0 reward will not affect a sum as calculated at the end of a task, but it may cause instability when using TD methods which update in an immediate, local fashion.

There is still value in applying a transformation to the reward before doing a backup for any given task. One can eliminate some cases where recursive optimality does not imply hierarchical optimality, as outlined in Dietterich (2000a). Additionally, by increasing the reward for successful termination, it is possible to allow a greedy policy to guarantee convergence to an optimal policy in some cases in which it may otherwise get stuck in a local minimum.

Additionally, as identified by Ryan (2004b), hierarchies can be constructed such that a subset of the state space is never explored within a given task if supertasks are never acceptable in those states. In the ordinary implementation of OPHRL, commitment can be required to learn even a recursively optimal policy. Therefore, choosing to continue a task to completion, even if supertasks no longer support the execution of the task, may be warranted.

### 3.3 Gated Temporal Second Difference Traces

As described in section 2, it is important to avoid incorporating rewards from exploration when attempting to learn off-policy. As it turns out, TSDT is a temporal difference method almost ideally suited to operating under this limitation. However, it can benefit from further modification.

One key observation that grants TSDT its power is that actions that appear to be exploratory when they're taken may later turn out to be the best choice. A non-greedy choice can turn out to be quite good, enabling the flow of information to be turned on. The same issue appears in the detection of exploration in subtasks. In fact, TSDT suffers from the problem that an action appearing to be optimal in a subtask may later turn out to be suboptimal. An entry can persist in the trace after it turns out that the subtask was exploring at that time. This problem will disappear as subtasks converge to their optimal value functions, but this could pose a serious problem for non-episodic or long-running tasks.

The gated temporal second difference trace (GTSDT) can resolve this issue by storing the information necessary to reassess the optimality of decisions taken in subtasks in the trace. Rather than excluding entries from the trace entirely when subtasks are behaving suboptimally, all entries are stored in the trace so that they may be allowed to update whenever subtasks appear to have behaved optimally. Entries in a trace can potentially become blocked and unblocked many times before the estimated value functions for the corresponding subtasks converge on their true value functions.

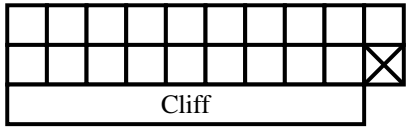

Figure 4: A shorter 10x2 cliff-walking domain.

## 4 Experimental Results

### 4.1 Cliff-Walking

We examine a 100x2 cliff-walking domain–a longer version of the domain depicted in figure 4. Four deterministic move actions can be attempted from each of the 199 non-terminal states. All actions result in a reward of $-1$ except for the terminal actions, which yield 200 for success and $-200$ for failure.

**Agents** A flat reinforcement learning agent simply decides between the four move actions from each state. We have constructed a hierarchical agent, depicted in figure 5, which chooses between a subtask which attempts to solve the traditional cliff-walking task by getting to the bottom-right corner, and a subtask which attempts to terminate by jumping off the cliff. Given terminal rewards of 200 and $-200$, solving the traditional task is always preferred by an optimal policy. The problem then is for the agent to efficiently learn both how to solve the traditional cliff-walking task and that solving it is always preferable to jumping off the cliff.

All hierarchical agents have no commitment to completing subtasks and explore with a fixed epsilon-greedy strategy, $\varepsilon = 0.1$. The flat agent and all subtasks explore with Boltzmann exploration, $T = 0.5$. All agents use all-goals updating to speed learning.

**Results** Figure 6 demonstrates that all hierarchical agents perform strictly worse that the flat agent, as expected. Both Fixed Q(0) and GTSDT learn quite well, but Naive Q(0) does not converge. GTSDT is able to learn more effectively than Fixed Q(0) primarily because Q-values corresponding to states far from the goal cannot be updated frequently given the lack of commitment to completing subtasks.

### 4.2 The Taxicab Domain

In the taxicab domain (Dietterich 1998), an agent is tasked with the problem of picking up a passenger and delivering him to his destination in as few steps as possible. The environment is a 5x5 grid world. There are four cells which serve as possible starting locations and possible destinations for the passenger. There is a refueling station near the middle of the map. Additionally, there are six impassable walls (or 26 counting the walls surrounding the map).

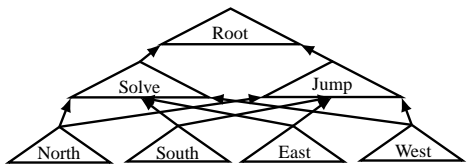

Figure 5: HRL cliff-walking agent.

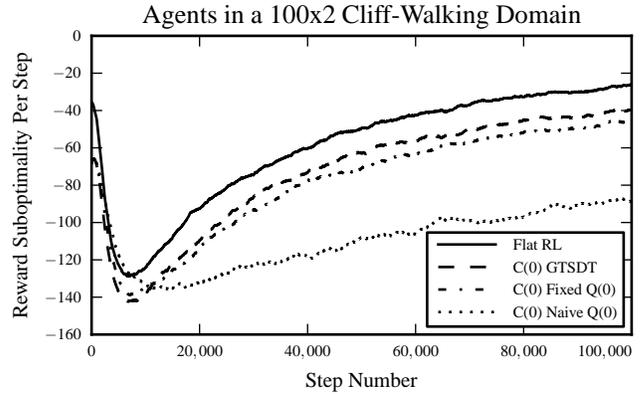

Figure 6: Online performance of agents exploring a 100x2 cliff-walking domain.

There are seven actions available to an agent at all times. Attempting to move north, south, east, or west automatically results in the taxi moving one cell in that direction unless there is a wall in the way, in which case the move action is ignored and the taxi remains in place. Fuel decreases by 1 unless the move action is ignored. Pickup always results in the passenger being picked up if the taxi does not have the passenger and is at the passenger's starting location. Putdown always results in the passenger being put down if the taxi has the passenger and is at the destination. Refuel always sets the amount of fuel to 12 if the taxi is at the refueling station.

Each of the seven actions takes 1 unit of time. Move, pickup, putdown, and refuel actions each yield a reward of $-1$ except in the following cases. Refuel, pickup, and putdown each yield a reward of $-10$ instead if the action is impossible when attempted. Move yields an additional reward of $-20$ if it causes fuel to drop below 0, resulting in failure of the trial. Putdown yields an additional reward of 20 if it causes the passenger to arrive at his destination, resulting in the successful termination of the trial.

**Agents** The hierarchy depicted in figure 8 is used with a few modifications from the version for HSMQ/MAXQ (Dietterich 1998). Q-values are shared between subtasks based on destinations. Rewards that are transformed to 0 in HSMQ are rejected instead. Terminal rewards for subtasks are transformed from $-1$ to $0$. Subtasks explore greedily. Finally, all tasks use all-states updating.

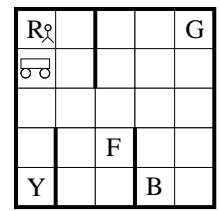

Figure 7: Taxicab Grid World Environment.

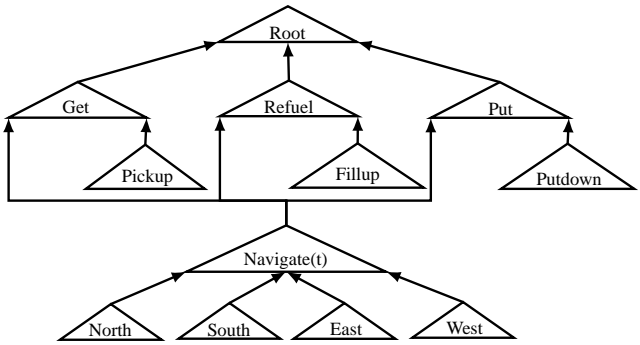

Figure 8: Hierarchical agent for the taxicab domain, as described in (Dietterich 1998).

**Results** Here we test fixed Q(0), fixed one-step intra-option learning (OSIO), and gated temporal second difference traces (GTSDT) while exploring with full commitment to completing subtasks. Additionally, all three algorithms are tested with a reduction in commitment from 1 to 0 over course of the 100,000 episodes. In the latter case, the cooling rate for Boltzmann exploration is increased from 0.999947 to 0.999924.

In terms of the policies learned after 100,000 steps, GTSDT does better than Fixed Q(0) which does better than Fixed OSIO, regardless of the level of commitment to completing subtasks. In terms of online performance, depicted in figure 9, GTSDT is always on top, but Fixed OSIO does better than Fixed Q(0) in terms of online performance if commitment is reduced significantly.

Linearly reducing commitment from 1 to 0 results in better policies for all three algorithms. Furthermore, online performance improves for both GTSDT and Fixed OSIO while only slightly decreasing online performance for Fixed Q(0). Fixed Q(0) and Fixed OSIO with reduction in commitment are omitted from figure 9 for space reasons.

## 5 Discussion and Future Directions

We identified a significant difficulty in attempting to learn off-policy in hierarchical learning systems. Solutions for Q-learning and Watkins' Q($\lambda$) are not ideal, requiring the discarding of potential learning, but one-step intra-option learning and temporal second difference traces handle the problem more gracefully.

We have demonstrated that it is possible for hierarchical reinforcement learning systems to learn efficiently without a commitment constraint, contrary to a claim in the literature that commitment should be critical for efficient learning. Furthermore, we demonstrated that reduction in commitment can actually help temporal difference methods learn more quickly.

The approach we explored for reducing commitment is somewhat ad hoc. It would be interesting to investigate more sophisticated exploration strategies capable of deciding whether or not commitment is valuable at any given time, as opposed to assuming that commitment is most valuable at the start of exploration.

## Acknowledgments

I would like to thank Professor John Laird and the University of Michigan for their support. I would like to thank the Soar Group, including Professor John Laird, Jon Voigt, Nate Derbinsky, Nick Gorski, Justin Li, Bob Marinier, Shiwali Mohan, Miller Tinkerhess, Yongjia Wang, Sam Wintermute, Joseph Xu, and Mark Yong for helping me to refine the presentation of these ideas. Additionally, I would like to thank Professor Satinder Singh for meeting with me to discuss some of his past work.

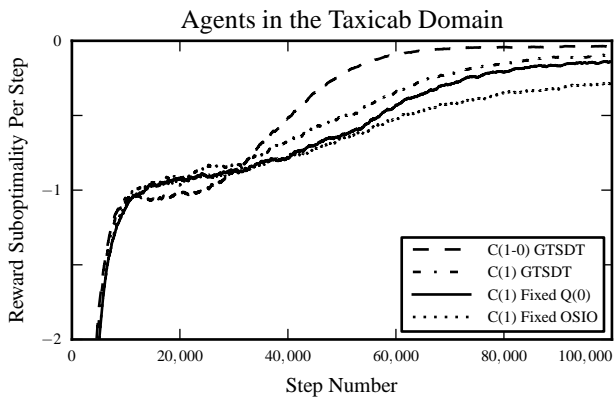

Figure 9: Online performance of agents exploring the taxicab domain.